\documentclass[journal]{IEEEtran}

\usepackage{times}
\usepackage{epsfig}
\usepackage{graphicx}
\usepackage{amsmath}
\usepackage{amssymb}
\usepackage[inline]{enumitem}
\usepackage{booktabs}

\begin{document}
\title{A Framework for Event-based Computer Vision on a Mobile Device}

\author{Gregor Lenz,
        Serge Picaud,
        Sio-Hoi Ieng
}

\markboth{}%
{Shell \MakeLowercase{\textit{et al.}}: Bare Demo of IEEEtran.cls for IEEE Journals}

\maketitle

\begin{abstract}
We present the first publicly available Android framework to stream data from an event camera directly to a mobile phone. 
Today's mobile devices handle a wider range of workloads than ever before and they incorporate a growing gamut of sensors that make devices smarter, more user friendly and secure. Conventional cameras in particular play a central role in such tasks, but they cannot record continuously, as the amount of redundant information recorded is costly to process. 
Bio-inspired event cameras on the other hand only record changes in a visual scene and have shown promising low-power applications that specifically suit mobile tasks such as face detection, gesture recognition or gaze tracking. 
Our prototype device is the first step towards embedding such an event camera into a battery-powered handheld device. 
The mobile framework allows us to stream events in real-time and opens up the possibilities for always-on and on-demand sensing on mobile phones. To liaise the asynchronous event camera output with synchronous von Neumann hardware, we look at how buffering events and processing them in batches can benefit mobile applications.
We evaluate our framework in terms of latency and throughput and show examples of computer vision tasks that involve both event-by-event and pre-trained neural network methods for gesture recognition, aperture robust optical flow and grey-level image reconstruction from events.
The code is available at https://github.com/neuromorphic-paris/frog
\end{abstract}


\IEEEpeerreviewmaketitle

\section{Introduction}
Mobile handheld devices are indispensable technology nowadays. 
An increasing range of their functionality is powered by machine learning models and in particular neural networks that are trained offline and deployed for inference.
To be able to perform on-device inference instead of computing in the cloud is important for a number of reasons~\cite{tensorflow2021guide}: 
\begin{itemize}
    \item Since there is no round-trip to a server, latency is greatly reduced.
    \item No data needs to leave the device, which avoids issues regarding the user's privacy.
    \item An internet connection is not required, which is beneficial to autonomy.
    \item Network connections are power hungry and should therefore be avoided if possible.
\end{itemize}

The number of parameters in neural network models grows rapidly every year. To be able to employ them on mobile devices that have constraint power budgets, we have seen the emergence of specialised neural network accelerator hardware and different approaches to reduce model size, number of floating point operations as well as latency~\cite{tan2019efficientnet, hinton2015distilling, jacob2018quantization}. 
But no matter the optimisations performed, neural networks still have to crunch a lot of redundant data, preventing mobile devices from continuously making use of them. 
For computer vision applications, this is because normally the visual scene is recorded using a conventional camera with a fixed frame rate, which is independent of the scene being recorded. 
Multiple cameras that are embedded into phones today not only serve to take pictures, but also facilitate tasks such as recognition of faces, gestures, objects, activities, and landmarks. 
Since image capturing and neural network inference are expensive, these tasks are often triggered by less computationally demanding sensors such as accelerometers or gyroscopes instead.
In today's systems with a tight power budget it is essential to intelligently manage high fidelity sensors and processing to reduce power consumption as much as possible.
But this can lead to latency issues or inaccurate triggering of the demanding processing in question and therefore consumes energy that could otherwise be saved. 

\begin{figure}[t!]
    \centering
    \includegraphics[width=0.48\columnwidth]{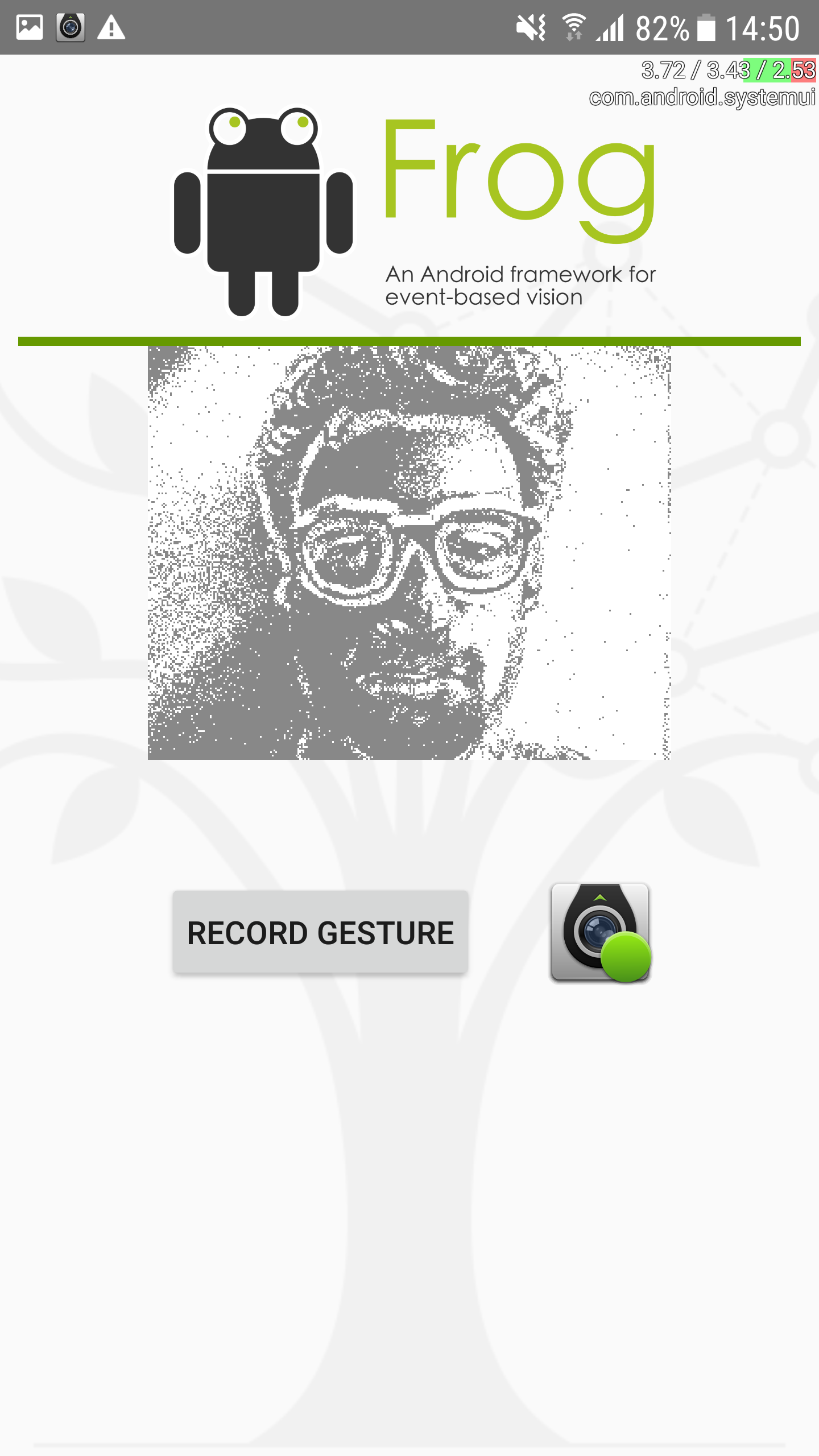}
    \includegraphics[width=0.48\columnwidth]{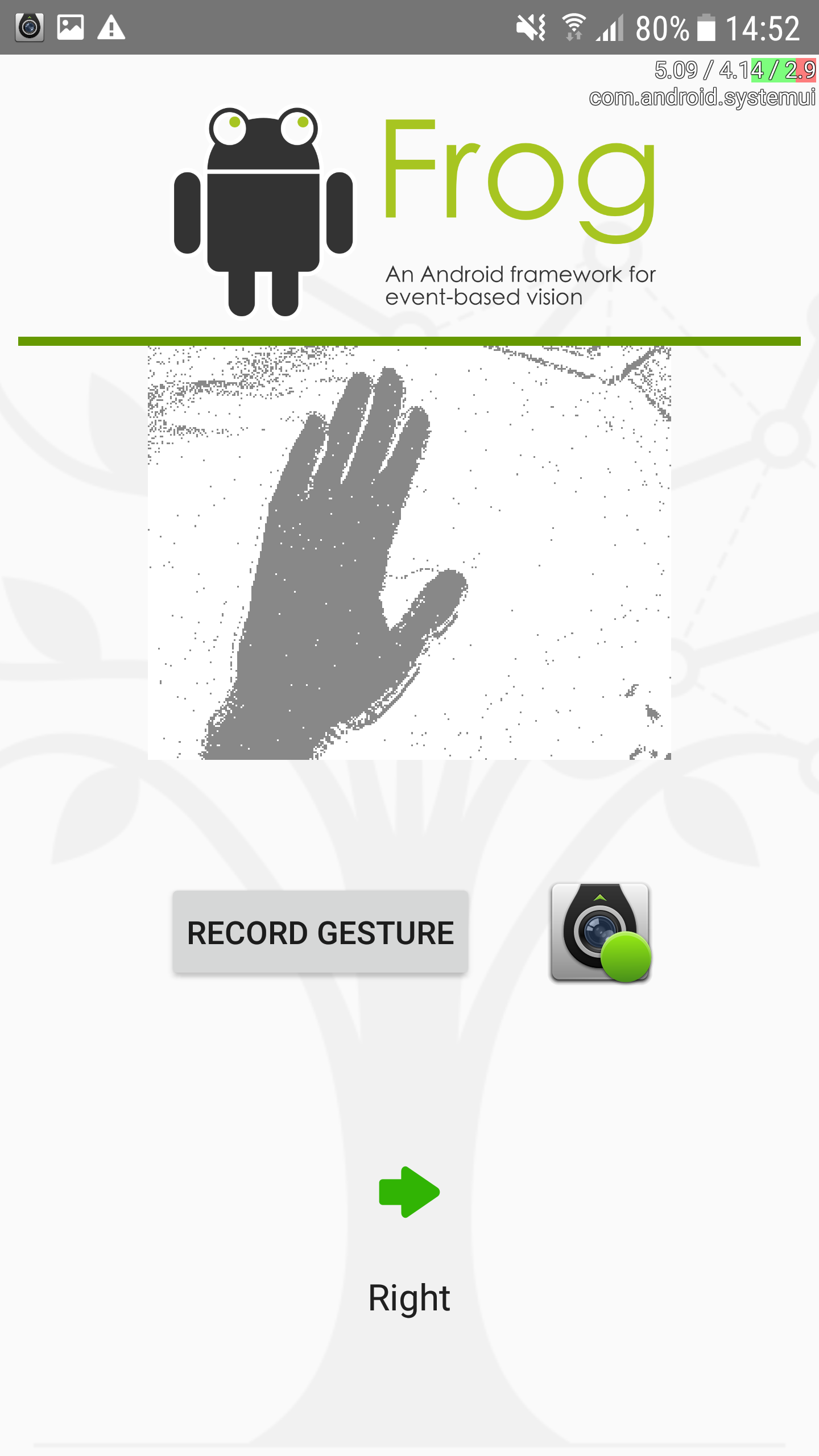}
    \caption{Screenshots of our Android app. \textbf{Left:} showing the live view of a connected event camera that renders in real-time. \textbf{Right:} Capturing a hand gesture being performed that signifies \emph{Right}.}
    \label{fig:screenshot}
\end{figure}

Event-based computer vision tackles the need for efficiency by using a novel image sensor. It employs event cameras~\cite{lichtsteiner2008128, posch2010qvga, serrano2013128, berner2013240, brandli2014240}, which are emerging, biologically-inspired vision sensors that can operate in an always-on fashion using very little power. Their pixels are fully asynchronous and only ever triggered by a change in log illuminance. The amount of events that are output is thus directly driven by activity in the visual scene and can range from a few to hundreds of thousands events per second. Power consumption is coupled to the amount of events recorded, which gives event cameras an edge for applications that might happen infrequently over time. 

Previous work has shown that mobile devices can profit from event cameras for low-power tasks such as visual activity detection~\cite{savran2018energy}, face detection~\cite{lenz2020event}, gesture recognition~\cite{maro2019event,maro2020event}, sensor fusion~\cite{ceolini2019sensor} or image deblurring~\cite{pan2019bringing, jiang2020learning}. Event cameras have successfully been employed on robotic platforms, which have similar limited power constraints~\cite{galluppi2014event, mueggler2015continuous, censi2014low, vidal2018ultimate}. Apart from the low power consumption, applications can also profit from high temporal resolution and good low-level light capture. 

\begin{figure}[ht]
    \centering
    \includegraphics[width=0.9\columnwidth]{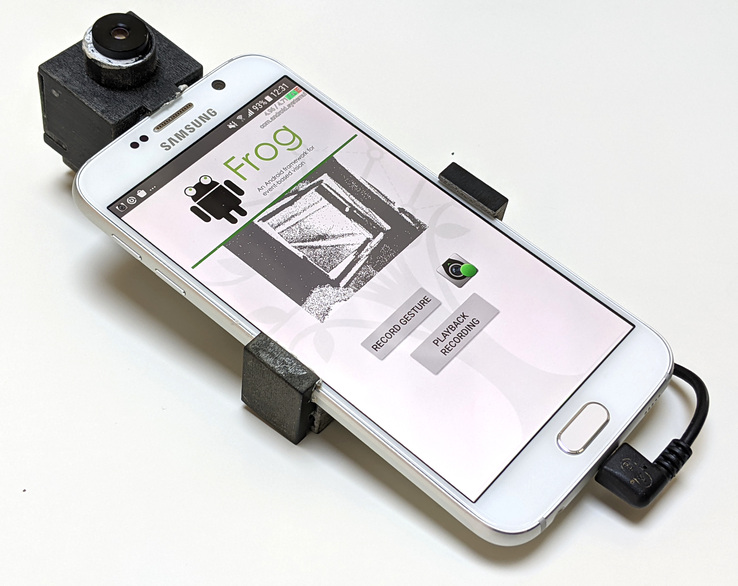}
    \caption{Prototype device, consisting of a Samsung Galaxy S6 and a small form-factor ATIS connected via mini-USB port. The camera is held in place with a 3D-printed frame that attaches to the phone. }
    \label{fig:prototype-device}
\end{figure}

In this work we use a prototype device for our experiments  consisting of an off-the-shelf mobile phone to which we connect a small form-factor ATIS~\cite{posch2010qvga} event camera via mini-USB connection as shown in Figure~\ref{fig:prototype-device}. We mount the event camera on a printed frame such that it faces the user. The device is self-contained and does not need any external cabling. 
Once the camera is connected, our Android framework is able to stream data from the event camera in real-time, enabling on-device processing. We make it straightforward for a user to deploy their own code using Android's Native Development Kit (NDK) or to use pre-trained neural network models in combination with the Tensorflow Lite library.
We give details about app architecture and how the different modules within depend on each other. 
We also provide examples of computer vision tasks that show the applicability of event cameras on mobile devices and benchmark throughput as well as latency to motivate further exploration in that direction.
Overall, our contributions are as follows: 
\begin{itemize}
    \item A publicly available mobile framework to stream events from an event camera in real-time.
    \item Real-time application of event-driven algorithms or pre-trained neural networks on events on a mobile phone.
    \item A self-contained device that uses a variable trigger in the form of an event camera for always-on computer vision applications.
\end{itemize}

Processing event by event from our camera can potentially achieve the lowest latency, as new information is integrated as soon as it is available. This kind of processing which does not use conventional frames needs rethinking computer vision algorithms from the ground up and has seen promising applications in event stream classification or detection~\cite{lagorce2016hots, haessig2018sparse, sironi2018hats, lenz2020event}.
Since we execute event-based algorithms on conventional von Neumann as opposed to specialised neuromorphic hardware, the event-by-event approach of asynchronous input does come with an overhead when repeatedly updating a state up to hundreds of thousands of times per second. von Neumann hardware is designed to compute on bulks of memory, and not for fine-grained parallelism.
We therefore examine the effect of buffering events, to be able to process them in batches. Depending on the algorithm, this can alleviate some of the computational burden, but also incurs latency.
Buffering events means balancing a power/latency trade-off that depends on the number of input events per second.
On one end of the spectrum, an input event rate of hundreds of thousands events per second for an active visual scene and a buffer size of 1 is likely to overwhelm a device such as a mobile phone with many updates. On the other end, a large buffer size when there are only few input events will not  trigger any update at all.
Depending on the application, we show how an acceptable trade-off can look like, to bring event-based computer vision to power-constrained mobile devices.

\section{Mobile Device and Event Camera}
\label{hardware-setup}
\begin{figure}[ht]
    \centering
    \includegraphics[width=0.7\columnwidth]{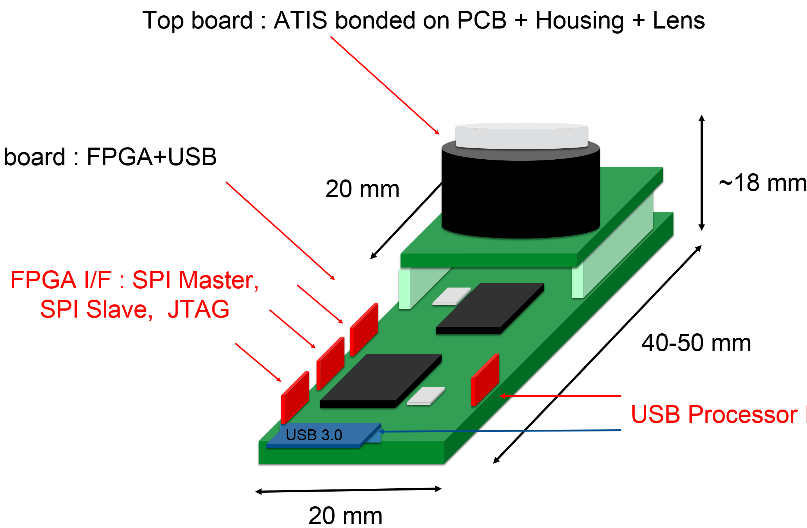}
    \caption{Small form-factor event camera assembly. the stacked printed circuit board is located within the housing on top of the phone, as shown in Figure \ref{fig:prototype-device}.}
    \label{fig:mini-atis}
\end{figure}

Our device prototype as shown in Figure~\ref{fig:prototype-device} consists of a Samsung Galaxy S6 smartphone and a small form-factor event camera. Small form-factor event cameras such as the embedded DVS~\cite{conradt2015board} have a lower spatial resolution and optimised power consumption in comparison to normal event cameras since they target battery-powered devices. Our low-power version of an ATIS~\cite{posch2010qvga} has a spatial resolution of $304 \times 240$ pixels, is fixed on a 3d-printed external frame and connected to the device via the mini USB port. 
The camera die of size $5000 \times 5000 \mu m^2$ with a fill factor of 30\% was fabricated using a UMC 180\,nm process. Power consumption for the chip depends directly on scene activity, where one pixel draws 300\,nW under static conditions or 900\,nW under high activity. The readout of events is facilitated using an FPGA and draws 30\,mW for high activity of all pixels. I/O communication for the USB connections needs further 20\,mW. 
The camera is embedded in a printed case on top of the mobile phone, to be able to directly face the user. In order to ensure flexibility and compactness, a stacked design of two printed circuit boards was chosen as depicted in Figure~\ref{fig:mini-atis}. In theory also other event cameras can be connected via USB as long as drivers are open source, although standalone cameras will need an external power supply. 

\section{Android Application Framework}

\begin{figure}[hbt]
\centering
\includegraphics[width=\columnwidth]{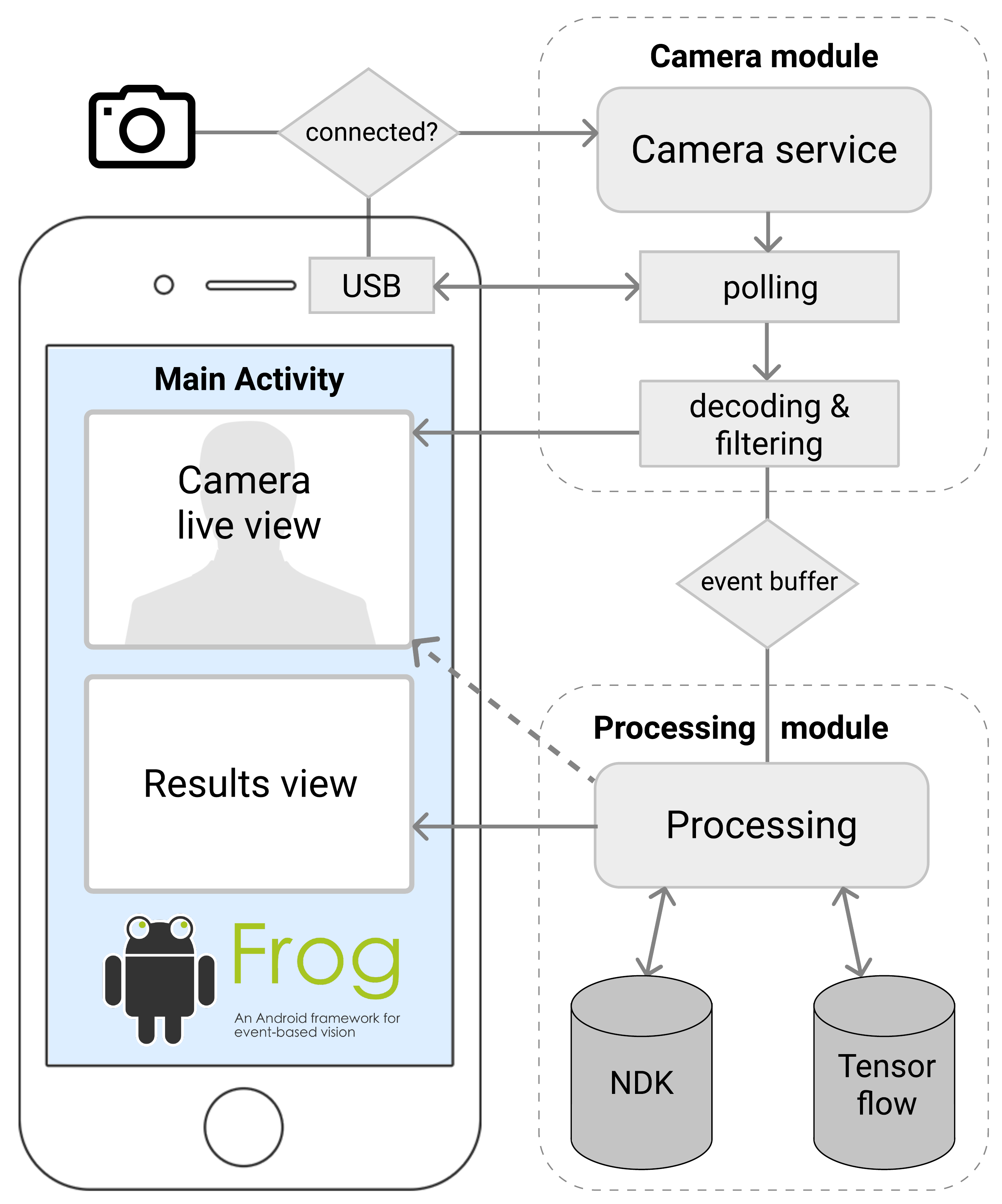}
\caption{Application software architecture. Based on Android, we make use of a \textbf{Camera module} to handle the streaming of events from a camera and a \textbf{Processing module} that is able to run different algorithms depending on the backend. Both modules update Views in the \textbf{Main Activity}.}
\label{fig:application-arch}
\end{figure}

Our proposed Android app facilitates the readout and processing of events from an event camera in a power-efficient way. We split the streaming of data packets via USB from the rendering and processing of user-defined code into separate modules, which are outlined in Figure~\ref{fig:application-arch}:
\begin{enumerate}
    \item The main activity which renders the user interface.
    \item The camera module that deals with transferring events from the camera as well as accumulating them in a buffer.
    \item The processing module that can be called on demand to execute algorithms on the events in the buffer.
\end{enumerate} 

From a functional standpoint, as soon as the event camera is connected, a live view will start rendering the camera output on the screen so that the user can have visual feedback of how they interact with the device as shown in Figure~\ref{fig:screenshot} on the left. Data received from the camera is checked for isolated noise events and constantly firing pixels, which are filtered and discarded to not unnecessarily strain the downstream processing. In this phase, there is no computationally heavy processing necessary. 

Whenever the user gives the signal to start processing the events with a pre-defined algorithm  by pressing a button on screen, the app accumulates events in an event buffer, where they await further processing. The live view continues uninterrupted. As soon as the buffer of a specific size is full, the processing routine will be called in a separate worker thread. 
The event buffer size can be adjusted, which indirectly determines how often the processing routine is called. If the buffer size is too small, the amount of computations per second might overwhelm the phone's CPU. If the buffer size is too large, results are presented to the main activity very infrequently and might impact user experience.  The right buffer size causes events to be processed in batches, balancing computational cost and result latency. 
The processing routine then returns a result depending on the algorithm used, such as a specific classification outcome or optical flow speeds. The result that is presented to the main activity can be displayed in a text box or used as an overlay for the live camera view. 
In the following part we will describe the 3 modules that the app consists of in more detail.

\subsection{Main Activity}
The main activity is responsible for the app life cycle and for rendering the user interface, bundling together the camera live view as well as results view. It is also responsible for handling the necessary permissions for USB devices, which a user has to agree to when they connect a new device. Any continuously ongoing processing such as USB polling has to be done in background threads, as otherwise the user experience would suffer if the interface started to lag or stall.
The live camera view is rendered at the native frame rate of the phone, which is 60\,Hz for the Samsung Galaxy S6. For efficiency reasons, the live camera view renders a binary bitmap at the native resolution of the event camera, which is then scaled up to display view size. The results view will update whenever the processing routine in the processing module returns a new result. 

\subsection{Camera Module and Event Buffer}
This module deals with receiving the events from the event camera via USB  and pre-processing them. As soon as such an event camera is connected, a \emph{camera service} as part of the main thread will be started, which deals with the camera initialisation and handles two background threads for \emph{polling} and \emph{decoding}. The event camera and its FPGA need 2-3 seconds to power up, after which the camera's biases are set and it is switched into readout mode. 
The \emph{polling} thread managed by the camera service is periodically querying the USB interface for new data packets, about every 1\,ms. A packet can be anything from 0 to 16\,kB, depending on the scene activity and therefore the event camera's output. Those packets are placed in a packet buffer, so that the polling can continue uninterrupted. 
The \emph{decoding} thread managed by the camera service takes a USB packet from the packet buffer whenever available, and converts the binary blob into a number of events. The user can decide to apply simple refractory periods for each pixel to prevent excessive firing of pixels, or to apply additional filtering to remove noisy events.
The same thread also directly updates the bitmap used by the Camera live view, at potentially much higher rate than the display refresh rate. 
If the user has triggered algorithm execution, the filtered events that were used to update the bitmap are then accumulated in an event buffer of size $N$. This buffer will act as a gate for downstream processing and will only trigger a computation when the buffer is full.

\subsection{Processing Module}
This module is responsible for the execution of algorithms using the batch of events that is passed from the event buffer. A third background thread is started whenever the processing routine is triggered. The routine can make use of different backends to make computation as efficient as possible. One option is the deployment of user code in C\texttt{++}, which can be executed natively on the phone using Android's Native Development Kit (NDK). The process routine can then call those native functions via the Java Native Interface, which take one or more events as parameter, to efficiently compute and return a result for that same batch. Calling a function through the Java Native Interface incurs an overhead, but the efficiency of native code execution often makes it worth wile. It should be noted that this backend uses a single background thread only.

Another option is to make use of a TensorFlow Lite backend~\cite{lee2019device}, which is a framework for neural network inference for edge devices with hardware support on Android platforms. A neural network that has been trained offline can be processed to suit the deployment on an Android phone, by fusing and dropping as many operations as possible or quantizing weights to reduce computational effort and latency. The compressed network can be bundled with the Android app. Given an input, such a condensed network returns a similar result as the full precision network up to an error margin.  A neural net that has been trained on events takes an accumulated event frame as input, so the event buffer size \emph{N} will typically be higher in this setting. The neural network output and result can also be reported to the main activity to update the result and/or the live view.

\section{Performance Measurement Methods}
We benchmark the components of our system that contribute to the overall latency from the point when the event camera emits an event until a result is computed and  measure the amount of events that can be handled per second. These components are the Camera module including its event buffer and the Processing module. 

\subsection{Camera Latency}
At first we want to measure how quickly we can transfer, decode and filter events sent from our event camera connected via USB. 
Depending on scene activity, the event camera can generate up to hundreds of thousands of events per second. We denote the rate of events/s recorded by the camera as $R$. It serves a proxy for activity in the visual scene. 
The latency incurred by the camera module is the time it takes to transfer, decode and filter a USB packet of events: $\lambda_\text{cam} = (t_\text{transfer} + t_\text{decoding} + t_\text{filter}) N_{packet}^{-1} $, where $N_{packet}$ is the number of events in a USB packet. The accumulated latency per second for the camera module is:
\begin{equation}
    \mathcal{L}_\text{cam}(R) = R \lambda_\text{cam}
\end{equation}

\subsection{Buffering Latency}
After the events have been decoded, they are placed in the event buffer.
Performing computation on each event individually incurs a large overhead when looking up and dereferencing functions~\cite{marcireau2020sepia}.
We can therefore accumulate multiple events in our event buffer of size $N$ to then process them as a batch. This typically saves overhead costs of creating new threads and updating states with every event, but the buffer size has to be chosen depending on the application and $R$. 
The accumulation of events causes latency per event that is the inverse of the input stream event rate, $\lambda_\text{buffer} = \text{R}^{-1}$. Bigger buffer sizes will cause longer latency and vice versa. The cost of moving data to and from the buffer is factored into camera and processing module respectively. To calculate the latency that is accumulated per second, we write:
\begin{equation}
    \mathcal{L}_\text{buffer}(R, N) = N \lambda_\text{buffer} s^{-1}
\end{equation}

\subsection{Execution Latency}
We measure the latency for a certain algorithm $A$ as a function of buffer size, $\lambda_\text{exec} = A (N)$. If this function exhibits sublinear behaviour, the algorithm benefits from batching operations. 
To calculate accumulated latency per second, we multiply by the number of executions per second:
\begin{equation}
    \mathcal{L}_\text{exec}(A, R, N) = \frac{R}{N} \lambda_\text{exec}
\end{equation}

Together, these terms provide us with a tool to measure latency:
\begin{equation}
    \mathcal{L}(A,R,N) = \mathcal{L}_\text{cam} + \mathcal{L}_\text{buffer} + \mathcal{L}_\text{exec}
\end{equation}

$\mathcal{L}(A,R,N)$ computes a dimensionless output that tells us how many seconds of latency is accumulated per second from the moment that an event originates at the camera to the point when an algorithm returns a result. Everything at or under a value of 1 will be able to run in real-time.

\section{Experiments and Results}
\label{sec:android-results}
We benchmark the amount of events that we can handle in real-time from our event camera within the camera module, calculate buffer latency for different input event rates $R$ and implement 3 different computer vision algorithms on a mobile phone with the help of our framework. 
In the following experiments, we study event buffer size and its effect on latency for gesture recognition, computation of optical flow and image reconstruction from events. 
Optical flow computation is a relatively lightweight algorithm, which gives low-level information about direction and speed of events and which can benefit from batching operations.
With gesture recognition we want to exploit the event camera's natural suitability as a motion detector to extract higher-level information and make use of an event-based learned model.
The frame reconstruction is an example of a comparatively inexpensive neural network model that has been trained on events directly and that can make use of the TensorFlow backend. It also serves as a connection between purely event-based and conventional machine learning applications.

\subsection{Measuring Throughput of Camera Module and Event Buffer Latency}
\label{results-latency}
For a scene where a user is holding the phone in front of them, we observe 0.91\,ms of latency on average to transfer a USB packet that encodes 1024 events.
The decoding of such a packet including filtering and setting the shared bitmap live view takes another 0.73\,ms on a separate thread on average. 
The filtering is done to alleviate computational burden on our test device, where we remove about two thirds of events from the input stream. For that we use a refractory period of 1\,ms, a spatiotemporal filter of 1 pixel and $1$\,ms and also remove 2 constantly firing pixels completely.
This results in 
\begin{equation}
    \lambda_\text{cam}  =  1.6 \pm 0.3 \mu s / Event
\end{equation}
of latency for transferring, decoding and filtering events from the camera. This translates to a maximum event rate of 624.39~\text{\,kEvents/s} that we can sustain for real-time live view using a single CPU thread. 

For reasons of reproducibility and comparability, we benchmark all downstream components using a pre-recorded dataset.  We use gesture recordings from the Navgesture database~\cite{maro2020event} which have been acquired with the same camera as ours. It contains 1342 recordings of 6 different mid-air hand gestures performed. About 1 billion events are distributed over 47 minutes of recording time, which when distributed equally corresponds to an average $R$ of 365.6\,kEvents/s. Applying the same filtering as in the previous camera module experiment, this leaves us with an average input $R$ of 113.9\,kEvents/s.


\begin{figure}[htb]
    \centering
    \includegraphics[width=\columnwidth]{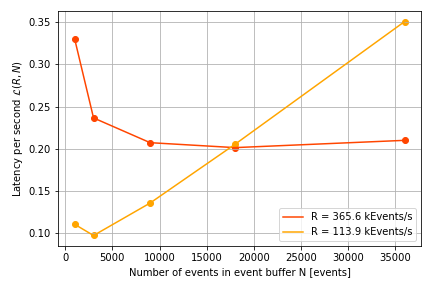}
    \caption{Accumulated latency per second for different event rates when computing event-based, aperture robust optical flow~\cite{akolkar2020real}. For high event rates (orange line) the overhead of calling a function repeatedly when the buffer size is low dominates the overall latency. For lower event rates (yellow line), buffer size can be considerably lower while still being able to compute in real-time. High buffer size combined with fewer input events means that events are spending a lot of time in the buffer which increases latency again. }
    \label{fig:optical-flow-latency}
\end{figure}

\subsection{Aperture Robust Event-based Optical Flow}
We implement and benchmark event-based aperture-robust flow as in Alkolkar et al~\cite{akolkar2020real}. Standard event-based flow as in Benosman et al.~\cite{benosman2012asynchronous} provides directions that are perpendicular to the surface formed by the events, which does not necessarily correspond to the true direction of motion. 
Alkolkar et al. propose  an algorithm that corrects the optical flow over a spatial region. It  can  be  divided  into  three steps: At first, local optical flow for each event is computed via a least-squares minimisation of a plane, as described in \cite{benosman2013event}. In the second step, different spatial scales are evaluated over which the mean magnitude of local flows is maximised. In the third step, the mean direction of local flows is calculated for the previously found optimal spatial scale.

\begin{figure}[htb]
    \centering
    \includegraphics[width=\columnwidth]{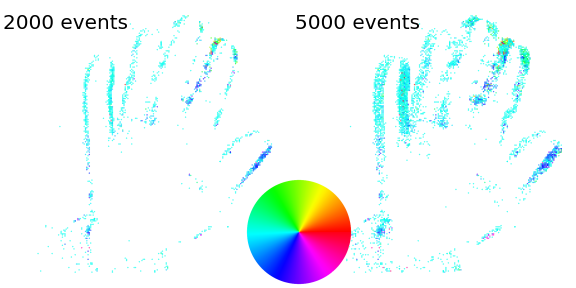}
    \caption{Aperture-robust event-based optical flow~\cite{akolkar2020real} computed on a recording of a person performing a mid-air hand gesture. The events are colour-coded to represent the direction of computed flow. \textbf{Left:} 2000 events are taken into account when computing flow, which provides a thin outline but correctly detects the direction. \textbf{Right:} 5000 events are accumulated for visualisation, equally achieving good results in terms of direction sensitivity. The motion looks blurry due to the longer time window of events.}
    \label{fig:optical-flow-comparison}
\end{figure}

This algorithm, especially the second step, can directly benefit from batching operations, as multiple spatial scales can be evaluated more efficiently. 
Figure~\ref{fig:optical-flow-latency} shows the effect of accumulated latency per second when computing the corrected flow. The algorithm computes in real-time even for high event rates of $>365$\,kEvents/s. We observe a drop in accumulated latency for batch sizes at around 5000. For larger batch sizes, the effect of buffer latency starts to dominate, which is especially apparent for lower input event rates. 
Independent of the buffer size, we achieve correct flow measurements that indicate the direction of the gesture performed, as shown in an example in Figure~\ref{fig:optical-flow-comparison}.

\begin{figure}[htb]
\centering
\includegraphics[width=0.95\columnwidth]{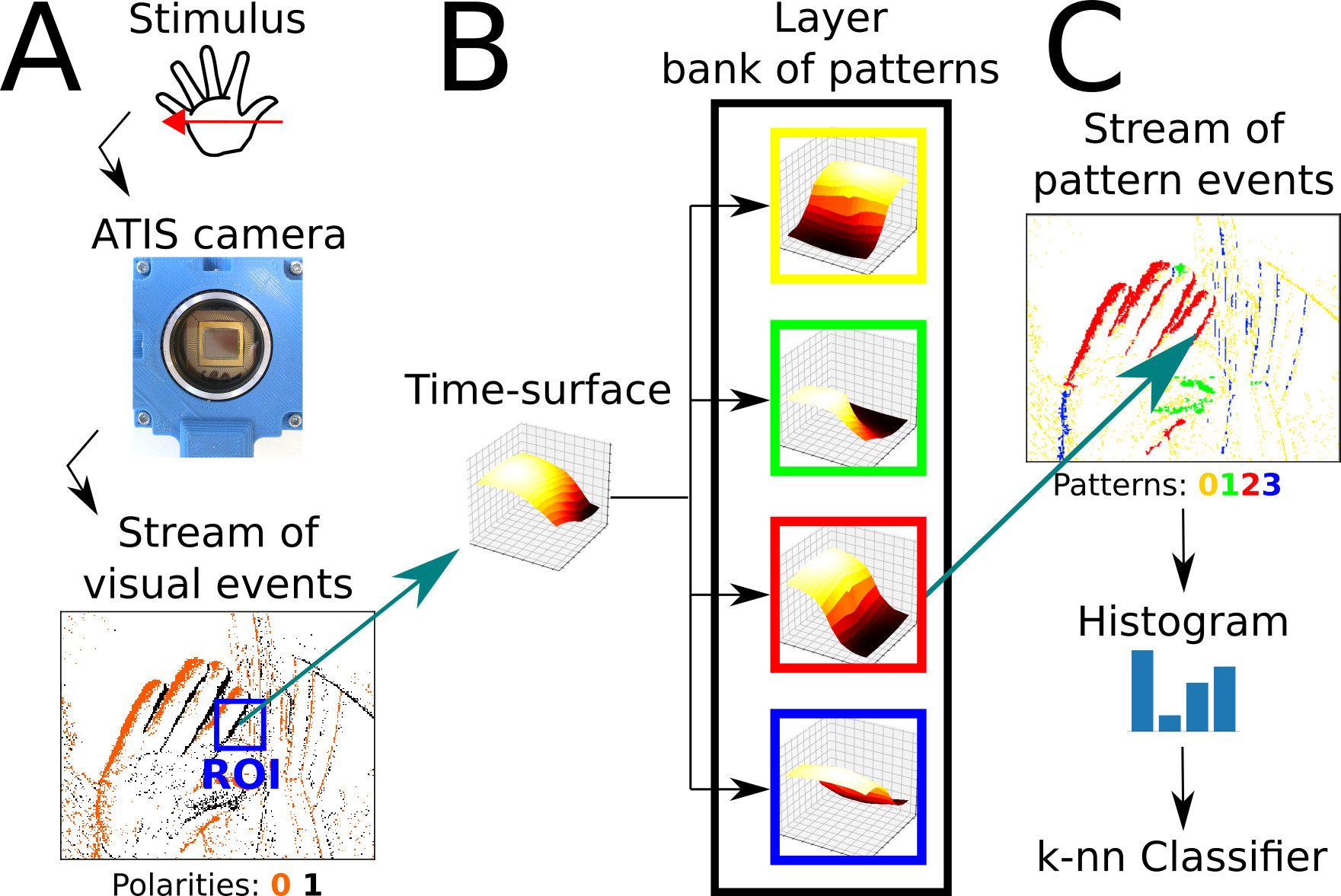}
\caption{(A) A stimulus is presented in front of a neuromorphic camera, which encodes it as a stream of events. (B) A time-surface can be extracted from this stream. (C) This time-surface is matched against known patterns, called prototypes. The number of occurrences of each prototype can be used as a feature for classification in the form of a histogram. Figure adapted from Maro et al.~\cite{maro2020event}}
\label{fig:gesture-recognition-summary}
\end{figure}

\subsection{Event-by-event Gesture Recognition}
\label{results-event-pipeline}

We implement and benchmark an event-based gesture recognition algorithm including background suppression for mobile phones~\cite{maro2020event}. The algorithm was trained using the NavGesture dataset~\cite{maro2020event} so that a user can perform one of 6 gestures: Up, Down, Left, Right, Select and Home.
An overview of the algorithm is shown in Figure~\ref{fig:gesture-recognition-summary}. As soon as the user presses a button, two seconds of events are recorded, at the end of which a predicted gesture is displayed in the result view.  
The processing happens in two stages. During the two seconds of input events, the algorithm computes a spatio-temporal descriptor called a \textit{time surface}~\cite{lagorce2016hots} for each event, which will be used as features during classification. The time surface represents the spatio-temporal context of an incoming  event by linearly decaying events in its surroundings and encodes both static information such as shape and dynamic information such as trajectory and speed.
The time surface is then matched against learned time surfaces prototypes using a HOTS architecture~\cite{lagorce2016hots}, triggering activation for the closest matching one.
This process happens continuously for the duration of the two seconds. 

\begin{figure}[htb]
    \centering
    \includegraphics[width=\columnwidth]{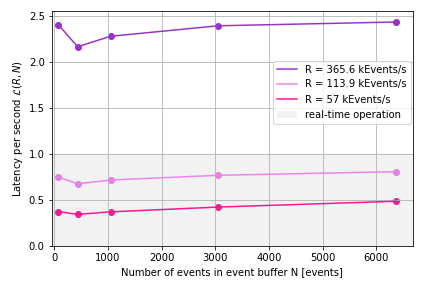}
    \caption{Accumulated latency per second when computing HOTS~\cite{lagorce2016hots} features for classification. Features can be generated in real-time for input event rates of about 150\,kEvents/s and beneath. From the measured stability in accumulated latency over batch size we can conclude that the algorithm, making use of a single thread and the NDK backend, does not benefit from batching.}
    \label{fig:latency-gestures}
\end{figure}

Figure~\ref{fig:latency-gestures} shows how the event buffer size impacts accumulated latency per second. This algorithm as currently implemented does not profit from batched operation, so we can see that latency is relatively stable. We do notice a slight overhead when buffer size approaches 1, and also see the impact of buffer latency $\mathcal{L}_\text{buffer}$ towards the other end. Overall, the feature generation can happen in real-time for event-rates at about $150$\,kEvents/s and less.

After the last feature has been generated, the second processing stage is triggered. 
The number of occurrences of all prototypes over a period of time is compiled into a histogram which is used as the gesture signature. The classification is done using k-Nearest-Neighbours. Here there is no option to break down or buffer the computation, so we just provide measurements of mean latency for the second processing stage of classification in Table~\ref{tab:classification-latency} when no event filtering is applied. The \emph{home} gesture with its dynamic back and forth motion causes many more events to be recorded, which has a significant impact on the time to prediction.

\begin{table}[htb]
    \centering
    \caption{Classification latency for 6 different gestures from the Navgesture database~\cite{maro2020event}. Mean latency is calculated over 5 trials each.}
    \resizebox{\columnwidth}{!}{%
    \begin{tabular}{ccccccc}
    \toprule
         & \textbf{Up} & \textbf{Down} & \textbf{Left} & \textbf{Right} & \textbf{Select} & \textbf{Home}  \\
    \cmidrule{2-7}
    \textbf{mean latency} [ms]  &  94.2 & 49 & 78.3  & 41 & 46.8 & 2825.6 \\
    \bottomrule
    \end{tabular}
    }
    \label{tab:classification-latency}
\end{table}

\begin{figure}[ht]
    \centering
    \includegraphics[width=\columnwidth]{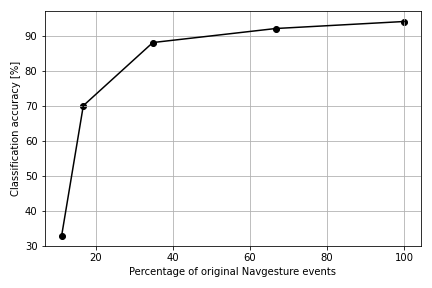}
    \caption{Event-by-event gesture classification results on NavGesture-sit~\cite{maro2020event}. By using spatiotemporal filters and refractory periods, we can reduce the amount of events and therefore computational cost considerably, without impacting classification accuracy too much. }
    \label{fig:gesture-classification-results}
\end{figure}

The filtering of input events has an impact on algorithm performance, so we plot the  classification accuracy over amount of filtered events in Figure~\ref{fig:gesture-classification-results}. 100\% of events correspond to all events from the Navgesture database. We show that we can filter about half of all events without a drastic drop in accuracy. 

\subsection{Leveraging Pre-trained Neural Networks for Image Reconstruction}
\label{results-neural-network-pipeline}
We convert the pre-trained model published by Scheerlinck et al. for fast image reconstruction from events~\cite{scheerlinck2020fast} to a TensorFlow Lite model that we can execute on the phone. This network has 38k parameters and uses voxel grids as input, which are accumulated and weighted accumulations of events into frames~\cite{gehrig2019end}. 
The aim is to show that we can reconstruct change detection events from the NavGesture database depending on scene activity, potentially allowing processing of conventional computer vision pipelines triggered by our inexpensive gate.
It might also serve as a way to render a visually appealing live view to the user.

\begin{figure}[ht]
    \centering
    \includegraphics[width=\columnwidth]{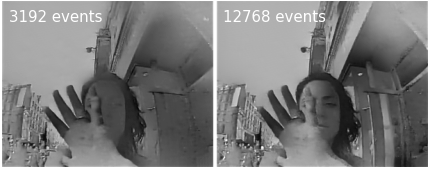}
    \caption{Gray-level frame reconstruction from events using a pre-trained FireNet~\cite{scheerlinck2020fast} model that has been converted to TensorFlow Lite. The two pictures differ in terms of number of events that have been used as input to the network. \textbf{Left:} 3192 events are used to create a voxel grid. The reconstructed frame exhibits strong smearing artifacts. \textbf{Right:} 12768 events are fed to the network, which increases latency, but also improves the visual results, for example details in the face or the door to the right.  }
    \label{fig:frame-reconstruction-comparison}
\end{figure}

Depending on the buffer size and unlike for the previous algorithms, we do observe results of different visual quality that depends on the event buffer size. Figure~\ref{fig:frame-reconstruction-comparison} shows the difference between the accumulation of 3192 events fed to the network that results in an image that exhibits strong smearing artefacts and lack of detail. The accumulation of 4 times the amount of events, 12768, results in much more consistent results. 

\begin{figure}[htb]
    \centering
    \includegraphics[width=\columnwidth]{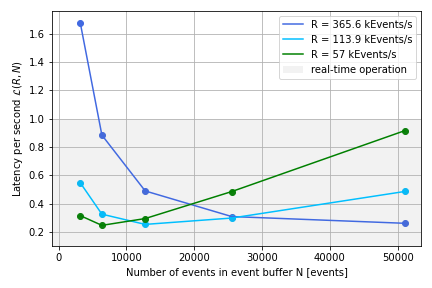}
    \caption{Latency per second $\mathcal{L} (R,N)$ over number of events in event buffer for gray-level frame reconstruction  from events using the FireNet~\cite{scheerlinck2020fast} neural network architecture and TensorFlow Lite backend. Using a low buffer size will trigger this expensive operation very often for high $R$, which is not permanently sustainable. If buffer size on the other hand is too big, no updates at all will be computed and events spend time waiting.}
    \label{fig:latency-frame-reconstruction}
\end{figure}

Voxel grids as a representation for events inherit some of the downsides of conventional frames, namely an abundant amount of redundant information. This is costly to process, and therefore we need to process events in higher buffer sizes. Figure~\ref{fig:latency-frame-reconstruction} shows the effect of buffer size on accumulated latency. Using the TensorFlow Lite backend, we can make use of special hardware acceleration, owing to the success of Deep Learning. It is therefore not a direct comparison to the previous two algorithms that make use of the NDK backend. 
A low buffer size of $<7000$ triggers the processing routine very often, swamping it with additional, redundant input data that is generated from the voxel grids. When $R$ is high, this is not sustainable for real-time operation, even with the use of dedicated hardware acceleration. A sweet spot seems to be at around a buffer size of 15000 events. 

\begin{figure}[ht]
    \centering
    \includegraphics[width=\columnwidth]{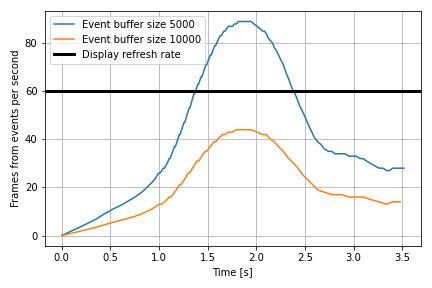}
    \caption{Event frames per second for an example gesture recording of 3.5\,s and two event buffer sizes of 5000 and 10000 events. The number of events and therefore frames depends directly on the visual scene activity and is thus highest when the gesture is performed. Display refresh rate is also shown for reference.}
    \label{fig:frame-reconstruction-frames-per-second}
\end{figure}

Until now we have worked under the assumption that events are evenly distributed over time, which is not the case for event camera recordings. 
Therefore we also want to show how many event voxel grids are generated over time for an example recording, shown in Figure~\ref{fig:frame-reconstruction-frames-per-second}. The bump in number of frames per second is caused by the gesture being performed, as the beginning and ending of the recording have little events that occur. Depending on the buffer size, this can trigger computation that refreshes the result more often than what the display is capable of rendering. Such computation could therefore be clipped to save energy. 

\section{Discussion}
This work presents a first step to integrate event cameras into mobile devices. 
Continuous processing of frames from conventional cameras is very costly on battery-powered devices and therefore only triggered when absolutely necessary. 
By swapping the conventional camera for our event camera that naturally acts as a motion detector, we can reduce computational load when there is no new information in the visual scene, and at the same time reduce latency to a minimum when computing results for fast motion. 
This approach is complementary to previous efforts of shrinking model sizes or algorithm foot prints.

We show that we can process input in real-time depending on different scenarios of visual activity.
The algorithms we tested are subject to a trade-off between computational demand and latency. 
It is worth mentioning that we process event-based data via the NDK backend on a single CPU thread on conventional von Neumann hardware, which is not designed for the level of fine-grained parallelism needed in some event-based approaches.
Nevertheless, our results on optical flow and gesture recognition that can be computed in real-time show the efficient nature of event-based computation.
The  TensorFlow Lite backend on the other hand makes use of special hardware acceleration such as the GPU to be able to reach sustainable throughput rates even though a lot of redundant information is generated in this case. 
In reality, our input event rate $R$ will change continually. Even if an algorithm accumulates more than a second of latency per second, computations can be skipped or allowed to catch up over time if input event rate drops again. 
 
To increase the efficiency even further, one option would be to dynamically adjust event buffer size so as to minimise $\mathcal{L}(A,R,N)$. Work in this direction was done by Tapia et al.~\cite{tapia2020asap}, although events are discarded if there are too many. 
Another option would be to cap computation at any rate higher than the display refresh rate. 
It would also be desirable to make use of a more efficient connection than USB to connect the event camera directly to the mainboard, such as Mobile Industry Processor Interface (MIPI) buses which are designed for low-power applications. 
Not only could the camera be integrated into the phone, but dedicated neuromorphic hardware which is specialised to execute spiking neural networks could help leverage the full potential of power-efficient computation. 
Event-based algorithms can make use of spiking neural networks that do not rely on the creation of voxel grids or other frame representations from events.

Connecting the world of event-based vision to a mobile device enables a range of potential applications such as face recognition, eye tracking, image deblurring, super slow-motion recordings or voice activity detection~\cite{savran2018energy}. 
Our work opens up the route to always-on sensing on battery-powered devices that make direct use of a vision sensor and that do not have to rely on low fidelity sensors to trigger expensive computation.
Neuromorphic cameras and algorithms can make current conventional systems more efficient by reacting to changes in the visual scene. The computation however is still done using von Neumann hardware. Even specialised neural network accelerators in the form of GPU-derived hardware do not exactly meet neuromorphic's demand for artificial neurons that can perform asynchronous computation. In the next chapter, we explore whether neuromorphic hardware can further reduce power costs to justify yet another piece of specialised hardware in the mix. 


\appendices
\section{}


\ifCLASSOPTIONcaptionsoff
  \newpage
\fi

\bibliographystyle{IEEEtran}
\bibliography{library}

%






\end{document}